%% file: main.tex
\documentclass[10pt,twocolumn,letterpaper]{article}

\usepackage[pagenumbers]{cvpr} 

\input{preamble}

\definecolor{mygray}{RGB}{127,127,127}
\definecolor{mygreen}{RGB}{93,174,86}

\usepackage{bm}
\usepackage{mathtools}
\usepackage{subcaption}
\usepackage{siunitx}
\usepackage[nopar]{lipsum}
\usepackage[export]{adjustbox}

\usepackage{pifont}
\usepackage{amsmath}
\usepackage{graphicx} 

\usepackage{bm}
\usepackage{mathtools}
\usepackage{subcaption}

\definecolor{mygray}{RGB}{127,127,127}
\definecolor{mygreen}{RGB}{93,174,86}

\makeatletter
\newcommand{\thickhline}{%
	\noalign {\ifnum 0=`}\fi \hrule height 1pt
	\futurelet \reserved@a \@xhline
}
\makeatother

\usepackage{caption}
\definecolor{cvprblue}{rgb}{0.21,0.49,0.74}
\usepackage[pagebackref,breaklinks,colorlinks,citecolor=cvprblue]{hyperref}
\usepackage{multicol,multirow}
\def\paperID{} 
\def\confName{CVPR}
\def\confYear{2024}
\newcommand{\ourmodel}{\textit{DrivingSphere}}
\title{\ourmodel: Building a High-fidelity 4D World for Closed-loop Simulation}

\author{Tianyi Yan$^{1,2}$, Dongming Wu$^{3}$, Wencheng Han$^{1}$, Junpeng Jiang$^{2}$, \\
Xia Zhou$^{2}$, Kun Zhan$^{2}$, Cheng-zhong Xu$^{1}$, Jianbing Shen$^{1}$\thanks{: Corresponding author:\textit{Jianbing Shen}}\\
$^{1}$SKL-IOTSC, Computer and Information Science, University of Macau
$^{2}$Li Auto Inc. \\
$^{3}$School of Computer Science, Beijing Institute of Technology\\
{\tt\small 
\{tianyi.yan123, wudongming97, wencheng256\}@gmail.com,jianbingshen@um.edu.mo}
}

\begin{document}
\maketitle

 

\input{sec/0_abstract}    
\input{sec/1_intro}

\input{sec/2_relatedwork}
\input{sec/3_method}

\input{sec/exp}

{
    \small
    \bibliographystyle{ieeenat_fullname}
    \bibliography{main}
}


\end{document}

%% file: preamble.tex
\usepackage[dvipsnames]{xcolor}


%% file: sec/0_abstract.tex
\begin{figure*}[ht]
    \centering
    \includegraphics[width=0.98\linewidth]{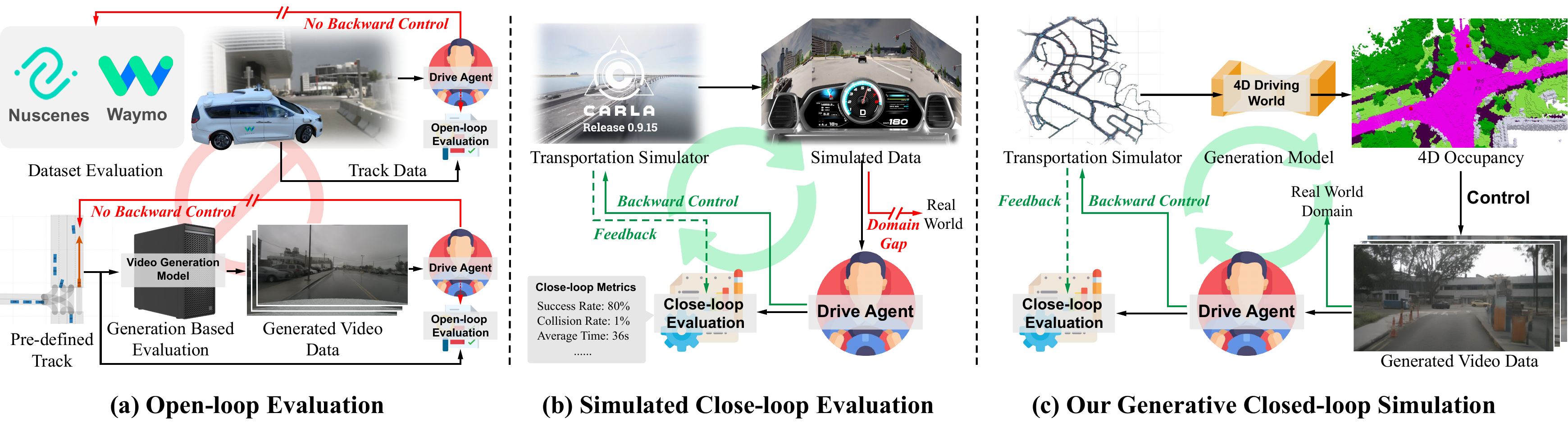}
    \caption{Comparison of frameworks for evaluating end-to-end Autonomous Driving (AD) algorithms. (a) \textbf{Open-loop evaluation} uses waypoint predictions along fixed routes in pre-collected datasets~\cite{Caesar_2020_CVPR_nuscenes}. Generative models~\cite{gao2023magicdrive,wang2023drivedreamer,zhao2024drivedreamer2} create diverse, realistic data but lack dynamic feedback to assess AD responses to dynamic changes. (b) \textbf{Simulated closed-loop evaluation}~\cite{dosovitskiy2017carla,li2021metadrive} offers feedback-driven, scalable environments where agent actions impact simulation dynamics; however, sensory outputs often differ from real-world data, limiting effectiveness for algorithms trained on real data. (c) Our \textbf{generative closed-loop simulation} framework, \ourmodel, addresses these limitations by delivering realistic visual inputs and continuous, responsive feedback between the AD agent and environment.
   }
    \label{fig:motivation}
    \vspace{-3mm}
\end{figure*}

\begin{abstract}
Autonomous driving evaluation requires simulation environments that closely replicate actual road conditions, including real-world sensory data and responsive feedback loops. However, many existing simulations need to predict waypoints along fixed routes on public datasets or synthetic photorealistic data, \ie, open-loop simulation usually lacks the ability to assess dynamic decision-making. 
While the recent efforts of closed-loop simulation offer feedback-driven environments, they cannot process visual sensor inputs or produce outputs that differ from real-world data. 
To address these challenges, we propose \ourmodel, a realistic and closed-loop simulation framework. 
Its core idea is to build 4D world representation and generate real-life and controllable driving scenarios. 
In specific, our framework includes a Dynamic Environment Composition module that constructs a detailed 4D driving world with a format of occupancy equipping with static backgrounds and dynamic objects, and a Visual Scene Synthesis module that transforms this data into high-fidelity, multi-view video outputs, ensuring spatial and temporal consistency. 
By providing a dynamic and realistic simulation environment, \ourmodel\ enables comprehensive testing and validation of autonomous driving algorithms, ultimately advancing the development of more reliable autonomous cars. The benchmark will be publicly released.
\url{https://yanty123.github.io/DrivingSphere/}

\end{abstract}

%% file: sec/1_intro.tex
\section{Introduction}

End-to-end autonomous driving algorithms \cite{hu2023uniad,jiang2023vadvectorizedscenerepresentationvad,chen2024vadv2,wang2023drivemlm,xu2024drivegpt4,shao2024lmdrive,zheng2024genad,chen2024end2end,li2024ego,han2024dme,chen2025asynchronous,zhang2024feedback,zhang2023cat} have made significant progress in recent years, making the accurate evaluation of these models a very urgent task.
For safe and responsible evaluation, it is essential to have a precise simulation environment that accurately reflects real-world driving conditions.
This requirement usually includes two aspects: firstly, the generation of high-fidelity sensory data, and secondly, the implementation of closed-loop feedback mechanisms.

The most common evaluation methods involve the forecasting of waypoints along fixed routes within collected public datasets~\cite{Caesar_2020_CVPR_nuscenes,sun2020scalabilitywaymo}, that is, \textit{open-loop} simulation approaches.
While these benchmarks offer realistic driving data, they have relatively fixed distributions and lack diversity~\cite{bevplanner}, limiting their ability to assess the generalization capabilities of autonomous driving algorithms. 
Even though several recent works~\cite{gao2023magicdrive,wang2023drivedreamer,zhao2024drivedreamer2} excel at producing diverse, photo-realistic data, continue to rely on the prediction of waypoints that are pre-established.
In summary, despite high-fidelity sensory data, these open-loop evaluation solutions cannot provide the dynamic feedback necessary for assessing how autonomous systems respond to dynamic changes and decision-making.

In contrast, closed-loop simulations~\cite{dosovitskiy2017carla,zhang2024lcsimlargescalecontrollabletraffic,wenl2023limsim,li2021metadrive} offer feedback-driven systems where an agent's actions influence and are influenced by other agents and the environment. 
For example, traffic flow-based simulation methods~\cite{zhang2024lcsimlargescalecontrollabletraffic,wenl2023limsim,fu2024limsim++} successfully enable multi-agent simulations. However, they lack the ability to process visual sensor inputs, limiting their interplay with vision-based end-to-end models. 
Game engine–based simulators~\cite{dosovitskiy2017carla,li2021metadrive} create scalable and physically realistic environments, but their outputs often diverge from real-world sensor data, restricting their utility in validating algorithms trained on actual inputs.

To address these challenges, we propose \ourmodel, a novel generative closed-loop simulation framework that leverages geometric-wise prior information to generate realistic and controllable driving scenarios.
In comparison with existing simulations, \ourmodel~has three distinguished features:
\ding{182} \textit{Rich simulation granularity.}
Unlike past methods that model only roads and cars~\cite{gao2023magicdrive,yang2024drivearena}, our approach allows for including previously unmodeled elements, such as buildings, vegetation, and other environmental structures. Although these non-traffic elements are not directly involved in traffic flow, their presence can significantly impact the input to driving models, thereby influencing decision-making processes in complex driving scenarios.
\ding{183} \textit{Physical and spatial realism.} 
Due to our model explicitly representing the scene and traffic actors in 4D space, it is capable of precisely depicting physical interactions and occlusion relationships across different traffic elements.
This ensures that each viewpoint and position naturally adheres to physical principles like depth and occlusion, thus achieving structured coordination of global road layout, traffic participants, and their behaviors.
\ding{184}  \textit{High visual consistency and fidelity.} 
Our model pays more effort to associate each traffic actor's appearance and unique ID within the scene, providing stable and high-fidelity temporal and spatial coherence across frames and views.

The proposed \ourmodel~contains two key modules: Dynamic Environment Composition and Visual Scene Synthesis. 
First, the Dynamic Environment Composition module aims to produce a coarse-grained 4D world with an occupancy grid format, which integrates large static backgrounds with dynamic traffic actors.
For this, we introduce a BEV-conditioned occupancy diffusion model, named OccDreamer, to generate static backgrounds.
It provides an infinite city-scale scene via progressive region extension.
Besides, we build an actor bank to manage key actors or objects, and update actors' spatial-temporal position.

Once the scene is generated, the second module, Visual Scene Synthesis,  converts it into high-fidelity, multi-view video outputs for AD system perception.
To fully capture the complexity of the 4D environment, we employ a dual-path condition encoding strategy. Specifically, the global branch, using a pretrained 4D encoder~\cite{wang2024occsora}, extracts geometric information and spatial-temporal relationships directly from the occupancy data, while the local projection branch produces view-specific semantic maps that align on a pixel-to-pixel basis, accurately capturing occlusions and depth variations.
To further ensure visual consistency, we design an ID-aware actor encoding mechanism that binds the appearance and identity information of traffic actors to their scene locations, which preserves spatial correlation across views and temporal consistency across frames.
With the custom-designed dual-path condition encoding and ID-aware actor encoding, 
the Visual Scene Synthesis generates spatially and temporally coherent video sequences. This results in a highly realistic simulation environment that is ideal for testing and validating AD systems.

By responsive interplay between AD algorithms and the simulated environment, \ourmodel~supports a closed-loop feedback loop, where actions from AD systems~\cite{hu2023uniad,jiang2023vadvectorizedscenerepresentationvad,chen2024vadv2,shao2024lmdrive} influence surrounding actors and scenes. This simulation platform closely mirrors real-world conditions, providing a reliable foundation for comprehensive AD validation.
Our extensive experiments show that \ourmodel~achieves superior performance in visual fidelity and temporal consistency, significantly reducing the domain gap between simulated and real-world environments. By excelling in both open-loop and closed-loop evaluations, \ourmodel~proves to be a reliable platform for dynamic agent-environment interplay in rigorous AD testing.
Ultimately, \ourmodel~will drive progress toward the development of safer and more reliable autonomous vehicles.
Overall, our contributions are four-fold:
\begin{itemize}
    \item We introduce \ourmodel, a novel geometry-aware closed-loop simulation framework that captures 2D visual and 3D geometric properties while seamlessly integrating with vision-based end-to-end driving agents.
    
    \item We propose the Dynamic Environment Composition that provides a comprehensive 4D driving world by modeling both static and dynamic elements, enabling intricate simulations across diverse environments.
    
    \item We design the Visual Scene Synthesis that enables long-term consistency and high fidelity, addressing challenges in generating continuous and multi-view video sequences for AD simulations.
    \item We introduce the Agent Coordination module for scalable multi-agent systems, enabling the integration of multiple vehicle agents and establishing a crucial feedback loop for validating autonomous driving algorithms.
\end{itemize}

%% file: sec/2_relatedwork.tex
\section{Related Work}
%
\noindent\textbf{Generative Models in Driving.}
Denoising Diffusion Probabilistic Models (DDPMs)~\cite{ho2020denoising} have transformed image generation by iteratively refining Gaussian noise into high-quality images. 
Based on this technique, generative models in autonomous driving have greatly advanced.
For example, BEVGen~\cite{swerdlow2024streetbevgen} and BEVControl~\cite{yang2023bevcontrol} use diffusion models to produce controllable street-view images from BEV layouts. 
More recently, diffusion-based video generation models~\cite{blattmann2023stablevideodiffusion,he2022latent} enable methods like~\cite{gao2023magicdrive,wang2023drivedreamer,zhao2024drivedreamer2,wen2023panaceapanoramiccontrollablevideo,wang2023drivingfuturemultiviewvisualdrivewm,gao2024vista} to generate multi-view driving videos, effectively supporting perception tasks.
Inspired by these 2D methods, advances in 3D scene generation have also emerged. Integrating diffusion processes into LiDAR generation improves the realism of LiDAR scenes~\cite{zyrianov2022learning,nakashima2024lidar,ran2024towardslidm,zyrianov2024lidardm,hu2025rangeldm}. Occupancy-based models have further progressed, with occupancy grids now central to 3D scene generation. SemCity~\cite{lee2024semcity} enhances outdoor 3D occupancy generation, while models like OccWorld~\cite{zheng2025occworld} and OccSora~\cite{wang2024occsora} use occupancy data for vehicle motion prediction and trajectory-conditioned 4D occupancy generation, respectively.
Unlike these, our OccDreamer combines text and BEV conditions to generate 3D occupancy data, dynamically enhanced with foreground motion and ego pose to form a complete 4D driving world. By conditioning on occupancy data, our VideoDreamer allows precise control over non-traffic elements and ensures spatial and temporal coherence, capturing essential physical relationships such as depth and occlusion.


\noindent\textbf{Autonomous Driving Simulations.} 
Autonomous driving simulation methods are developed from basic behavioral models to complex, interactive environments. 
Flow-based simulations~\cite{wenl2023limsim,fu2024limsim++,zhang2024lcsimlargescalecontrollabletraffic,feng2023trafficgen,ding2023realgen,tan2023lct} enable multi-agent interactions through long-term traffic planning.
However, these methods typically operate in 2D space and lack real-life visual outputs.
Platforms like CARLA~\cite{dosovitskiy2017carla}, SUMO~\cite{SUMO2018}, and MetaDrive~\cite{li2021metadrive} address these limitations by modeling traffic dynamics and supporting multi-agent interactions with visual contexts, 
but they have visual fidelity differences from real-world data, limiting their direct use with algorithms trained on such data.
While reconstruction-based methods~\cite{mildenhall2021nerf,kerbl20233dgs,ljungbergh2025neuroncap} and frameworks such as UniSim~\cite{yang2023unisim}, and MARS~\cite{wu2023mars} capture high-fidelity scenes, they are limited by the scope of the data used, restricting their ability to simulate varied and long-tail scenarios.
DriveArena~\cite{yang2024drivearena} recently advanced generative closed-loop simulation by using 2D traffic sketches to control visual outputs, integrating well with vision-based driving agents. However, reliance on 2D inputs constrains its modeling granularity and visual realism. Our approach overcomes these challenges by leveraging 4D occupancy data, generating visually coherent and geometrically accurate scenes that rigorously test driving agents in closed-loop simulations, equipping them to handle complex real-world conditions effectively.

%

%% file: sec/3_method.tex
\begin{figure*}[t]
    \centering
    \includegraphics[width=\linewidth]{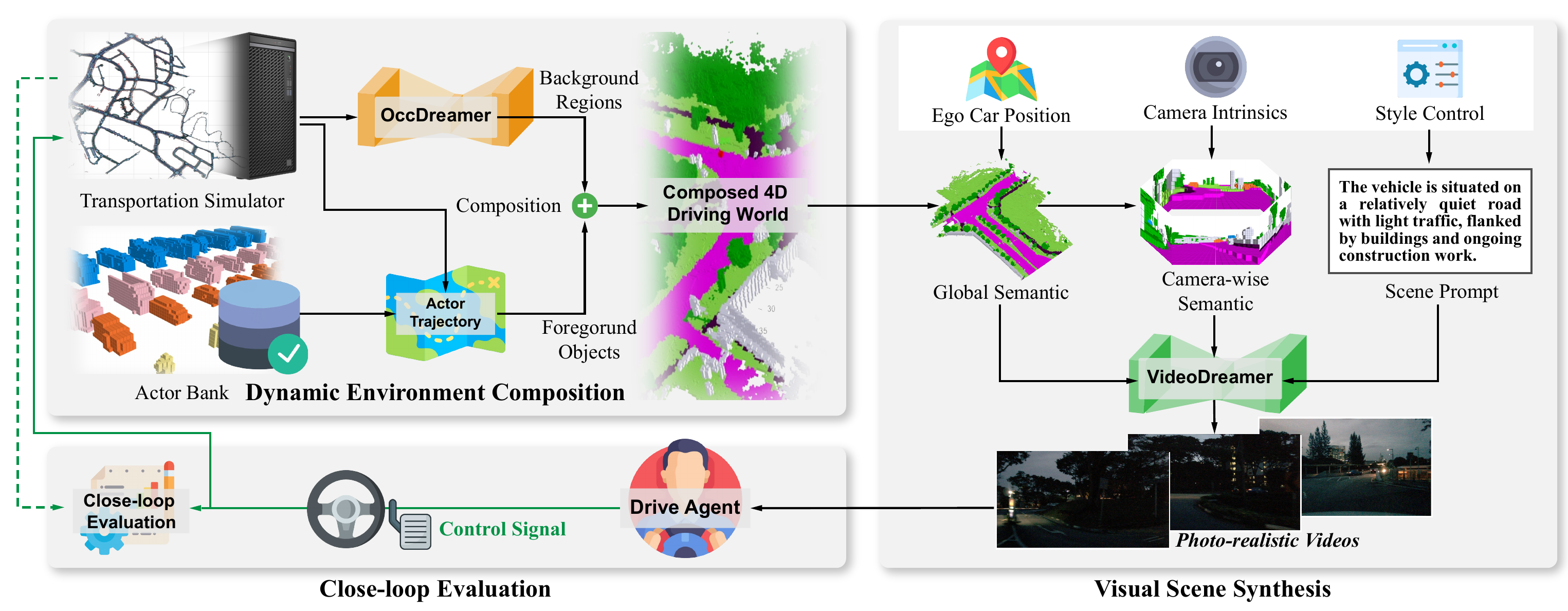}
    \caption{\textbf{Overview of \ourmodel~framework.} (a) The Dynamic Environment Composition module builds a 4D driving world, simulating real driving scenarios with backgrounds generated by OccDreamer, dynamic actors from an actor bank, and trajectories guided by a transportation simulator. (b) The Visual Scene Synthesis produces high-fidelity, photo-realistic video frames conditioned on global semantics, view-specific details, and scene prompts, supporting close-loop evaluation. Control signals enable adaptive feedback for driving agents, facilitating continuous testing and evaluation of driving algorithms in the simulated environment. }
    \label{fig:overview}
        \vspace{-3mm}
\end{figure*}
\section{Methodology}

\ourmodel~acts as a generative closed-loop simulation framework that combines occupancy-based 4D world modeling with advanced video rendering techniques, providing high-fidelity visual outputs that enhance both simulation realism and agent-environment interplay in autonomous driving scenarios. 
Specifically, \ourmodel~ begins with Dynamic Environment Composition (\cref{sec:occ}), which generates static background from a map sketch, selects traffic participants from the actor bank, and updates the actor location, to formulate a 4D driving environment with a format of occupancy.
Next, Visual Scene Synthesis (\cref{sec:video}) conditions the occupancy data surrounding the ego vehicle, accurately capturing occlusion relationships and fine-grained semantic information to generate high-fidelity multi-view videos. 
Finally, the closed-loop feedback mechanism (\cref{sec:close-loop}) enables dynamic, responsive adjustments, where autonomous agents continuously receive updated visual data and generate control signals that modify the simulation environment, providing a comprehensive platform for algorithm testing and refinement.
The overall framework is shown in \cref{fig:overview}.

\subsection{Dynamic Environment Composition}
\label{sec:occ}

Previous driving simulation methods have often overlooked static and varied elements like buildings, obstacles, and vegetation~\cite{dosovitskiy2017carla,yang2024drivearena}. Although these elements are not direct traffic participants, they are part of the perception input of AD systems,  influencing final driving decision-making. 
For instance, static objects like buildings may block sensors' line-of-sight, causing occlusions of other vehicles or pedestrians. 
Obstacles or vegetation can introduce sensor artifacts or false positives for AD systems. Both situations may result in incorrect decisions like unnecessary braking or lane changes. 
To address this issue, we propose a Dynamic Environment Composition to construct a comprehensive 4D Driving World equipping complex digital assets.

\noindent\textbf{4D Driving World Representation.}
Our 4D driving world representation includes three critical digital assets: \textit{static background scenes}, \textit{dynamic foreground actors}, and \textit{spatial position of these actors}.
Formally, we define this representation as $\bm{W}\! =\! \left\{ \bm{S}_\text{city}, (\bm{A}_1, \bm{P}_1), \dots, (\bm{A}_N, \bm{P}_N) \right\}_{n=1}^N$.
Here, $\bm{S}_\text{city}$ is a composite of multiple regional static scene $\left\{ \bm{S}_1, \dots, \bm{S}_K \right\}$. Each $\bm{S}_k \!\in \!\mathbb{R}^{H \times W \times D}$ is the $k^{th}$ static background scene that captures the spatial layout and static elements within the region. 
$\bm{A}_n \! \in \! \mathbb{R}^{4}$ is the $n^{th}$ actor of $N$ actors, such as vehicles and pedestrians, defined by a 3D coordinate and a semantic label. 
It is worth noticing that both the static scene $\bm{S}_k$ and dynamic actor $\bm{A}_n$ are represented as occupancy grids.
$\bm{P}_n  \!= \!\left\{ \bm{P}_n^0, \dots, \bm{P}_n^T \right\}_{t=0}^{T}$ describe actor position across $T$ frames, where $\bm{P}_n^t \in \mathbb{R}^{4}$ consists of a 3D relative position of the $n$-th actor at time $t$ with respect to the ego vehicle  and a yaw rotation $\theta$.
In the following, we will detail how to generate three assets.

\noindent\textbf{Static Scene Generation.}

\begin{figure}[t]
    \centering
    \includegraphics[width=\linewidth]{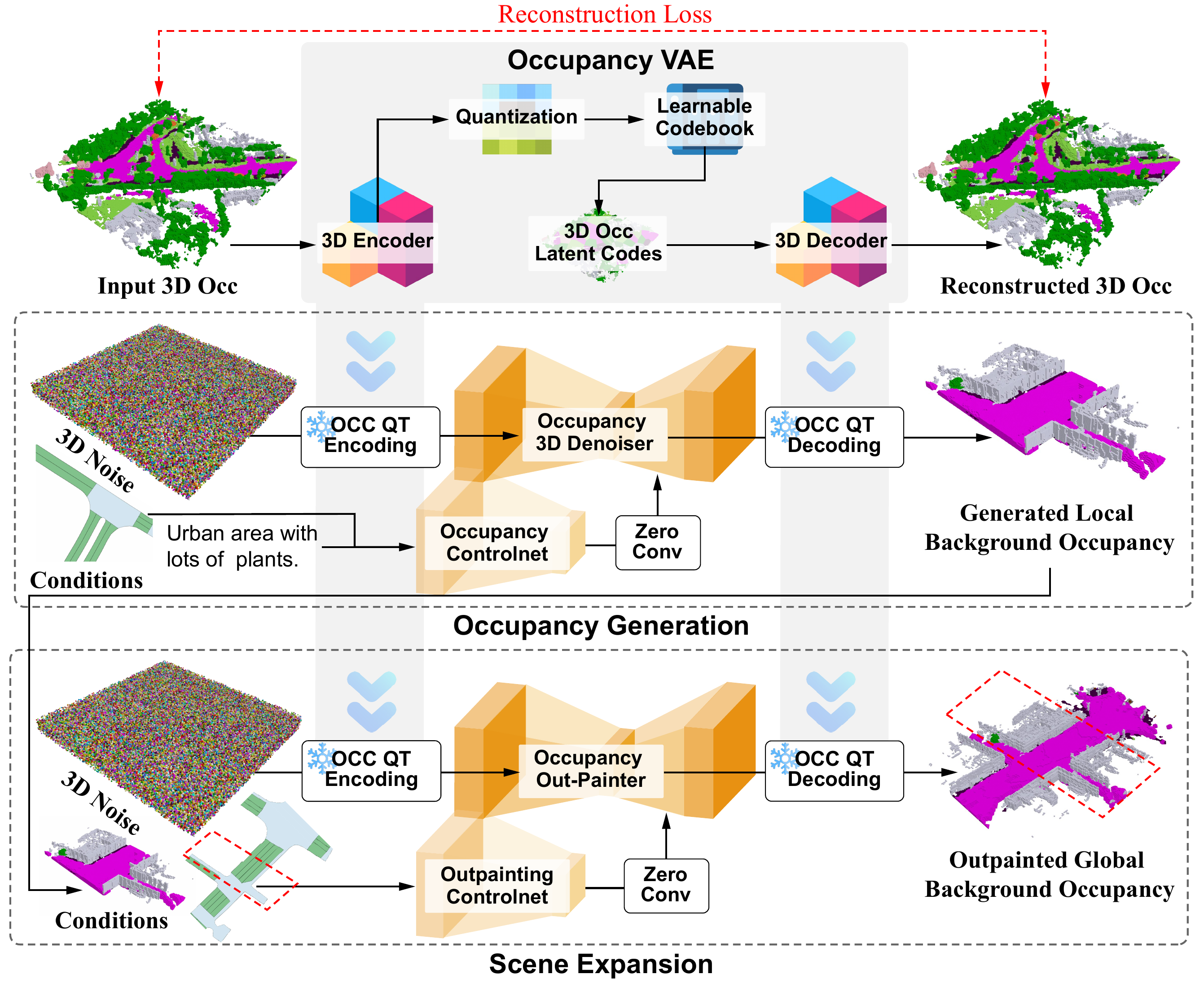}
    \caption{\textbf{The framework of OccDreamer}, which includes Occupancy Tokenizer, Region Occupancy Generation, and Scene Extension to generate a city-level background for 4D driving world.}
    \label{fig:occgen}
\end{figure}

A straightforward approach generating static 3D scenes $\bm{S}_\text{city}$ is directly using ground truth occupancy data from existing datasets, like the Boston area in nuScenes~\cite{Caesar_2020_CVPR_nuscenes}. However, this method is limited to specific regions captured during data collection, restricting its applicability to other urban areas. 
To address this limitation, we propose an occupancy diffusion model, OccDreamer, which is conditioned on BEV maps and textual descriptions, capable of generating static scenes for any desired urban region.
The framework of OccDreamer, which integrates the following components, is illustrated in~\cref{fig:occgen}.

Firstly, to efficiently train the diffusion model while addressing the computational complexity of processing 3D data, we adopt VQVAE~\cite{van2017neuralvavae,ho2020denoising,rombach2022highdiff} as an \textit{occupancy tokenizer}, mapping $\bm{S}_k$ to a latent feature $\bm{Z}^S$:
$\bm{Z}^S\!=\!\mathcal{F}^{occ}_{\text{VAE, Enc}}(\bm{S}_k)$.
The reconstructed scene is defined as $\bm{S'}_k\!=\!\mathcal{F}^{occ}_{\text{VAE, Dec}}(\bm{Z}^{S_k})$.
$\mathcal{F}^{occ}_{\text{VAE}}$ is trained with a combined loss $\mathcal{L}_{occ}$:
\begin{equation}
\vspace{-3pt}
    \mathcal{L}_{occ}=\mathcal{L}_{CE}(\bm{S}_k,\bm{S'}_k)+\alpha\mathcal{L}_{Lov}(\bm{S}_k,\bm{S'}_k)+\beta\mathcal{L}_{emb},
    \vspace{-1pt}
\end{equation}
where $\mathcal{L}_{CE}$ is the CE loss and $\mathcal{L}_{Lov}$ is the Lovász loss~\cite{berman2018lovasz}.
Secondly, considering varying road structures and complex regional topographies, we propose a controllable \textit{region occupancy generation module}.
It accepts a regional BEV map capturing the road structure and a language prompt describing abstract features  (like \textit{suburban areas with rich vegetation} or \textit{commercial districts with buildings lining both sides}) as input, and outputs a region occupancy $\bm{S}_k$.
Following the principles of diffusion models~\cite{ho2020denoising,zhang2023addingcontrolnet}, a CLIP encoder \cite{radford2021learningclip} converts the text prompt into an embedding $\bm{F}_\text{region}$. Then, $\bm{F}_\text{region}$ is injected into the denoiser $\epsilon_{\theta}^{s}$ via cross-attention.
Simultaneously, the BEV map is processed by a pre-trained image VAE~\cite{ho2020denoising} to extract the corresponding road embedding $\bm{F}_{M}$, serving as input for the ControlNet branch $\epsilon_{\phi}^{s}$. This enables precise control over the diffusion learning process in the latent space:
\begin{equation}
    \small
    \mathcal{L}_\text{occ}=\mathbb{E}_{\bm{F}_{M},\bm{F}_\text{region},\bm{Z}^{\bm{S}_k},\epsilon^{s},\tau} [\left \| \epsilon^{s} - \epsilon^{s}_\theta (\bm{Z}^{\bm{S}_k}_\tau,\bm{F}_\text{region}, \tau,\epsilon_{\phi}^{s}(\bm{F}_M)) \right \|^2],
    \label{eq:difscene}
\end{equation}
where $\bm{Z}^{\bm{S}_k}_\tau$ is the noisy version of $\bm{Z}^{\bm{S}_k}$ by adding $\tau$ steps noise $\epsilon^{s}$. After training, we use OccDreamer to create $\bm{Z}^\textbf{S}_k$ and decode it with $\mathcal{F}^\text{occ}_{\text{VAE},\text{Decode}}$ to obtain $\bm{S}_k$.

Thirdly, for spatial consistency across the entire 3D scene, we propose a \textit{Scene Expansion Mechanism} for building a coherent and city-level static scene $\bm{S}_\text{city}$.
Its core idea is to expand an initial region $\bm{S}_k$ to a neighboring region $\bm{S}_{k+1}$ by using the overlap between these regions as a conditioning constraint.
Formally, let $\bm{O}$ represent a binary mask that defines the overlapping area between two adjacent regions $\bm{S}_k$ and $\bm{S}_{k+1}$. This overlap serves as the conditional context for generating the next region, enforcing spatial coherence at the boundary. 
To generate $\bm{S}_{k+1}$, we first create a partially masked scene $\bm{S}_k^\text{partial}$ by applying $\bm{O}$ to $\bm{S}_k$: 
\begin{equation} 
\small
\vspace{-3pt}
\bm{S}_k^\text{partial} = \bm{S}_k \odot \bm{O}, 
\end{equation}
where $\odot$ denotes element-wise multiplication, extracting the overlapping section of $\bm{S}_k$.
We then employ a diffusion process to generate the neighboring region $\bm{S}_{k+1}$, conditioning on both the BEV map $\bm{M}_{k+1}$ (capturing the road structure of the target area) and the partially masked scene $\bm{S}_k^\text{partial}$. This diffusion process is represented as: 
\begin{equation} 
\small
\bm{Z}^{\bm{S}_{k+1}} = \mathcal{F}_\text{outpainting}(\epsilon^s, \bm{M}_{k+1}, \bm{S}_k^\text{partial}), 
\end{equation} 
where $\epsilon^s$ is the input noise, and $\mathcal{F}_\text{outpainting}$ denotes the diffusion model conditioned on the masked scene and BEV map.
After obtaining the latent representation $\bm{Z}^{\bm{S}_{k+1}}$, we decode it using the occupancy VAE decoder, resulting in the expanded region $\bm{S}_{k+1}$. The two regions are then combined into a larger scene $\bm{S}_\text{merged}$ through a merging operation: 
\begin{equation} 
\vspace{-3pt}
\small
\bm{S}_\text{merged} = \text{Merge}(\bm{S}_{k} \odot (1-\bm{O}), \bm{S}_{k+1}), 
\vspace{-1pt}
\end{equation} where $\text{Merge}$ ensures spatial continuity and logical consistency between adjacent regions.
The spatially continuous and logically consistent city scene $\bm{S}_\text{city}$ serves as an excellent digital background for the 4D driving world.

\noindent\textbf{Dynamic Actor Selection.} 
To complement static scenes, we populate the 4D driving world with dynamic actors, creating realistic traffic flows. An actor bank $\mathcal{B}$ is constructed, where each actor $\mathbf{A}_n \in \mathcal{B}$ consists of instance-level voxel data segmented from real or synthetic occupancy scenes, with unique geometric, semantic, and behavioral attributes, as well as a descriptive caption $\bm{L}_{\bm{A}_n}$ and an ID $I_{\bm{A}_n}$ to ensure consistency across frames.
Actors are selected based on semantic similarity to user-provided descriptions via CLIP, or randomly sampled from relevant categories if unspecified, ensuring contextual diversity.
This flexible selection process allows for dynamically relevant and diverse actor integration within the 4D driving world, supporting realistic and adaptable traffic simulation.

\noindent\textbf{Spatial Position of Actors.}
The generation of actor positions $\bm{P}$ is driven by the control signals output from autonomous agents (\cref{sec:close-loop}). Each actor is assigned an control signal set $\bm{C}_n = \left\{\bm{c}_n^0, \dots, \bm{c}_n^T\right\}$, where $\mathbf{c}_n^t$ represents the received control signal of actor $n$ at time $t$. These signals iteratively update the dynamic state of the world. For $n$-th actor, we have the position:
\begin{equation}
\small
\vspace{-3pt}
    \bm{P}^{t+1}_n=\bm{P}_n^t+\triangle \bm{P}^t_n(\bm{c}_n^t),
    \label{actorposition}
\vspace{-1pt}
\end{equation}
where $\triangle \bm{P}^t_n(\bm{c}_N^t)$ is the positional change resulting from the control signal $\bm{c}_N^t$ at time $t$.

\noindent\textbf{4D World Composition.}
With the static background, dynamic foreground actors and their positions calculated, we integrate them into the comprehensive 4D driving world. 
The world state at any time $t$ is represented by:
\begin{equation}
\small
\vspace{-3pt}
    \bm{W}^t=\pi (\bm{S},{(\bm{A}_n,\bm{c}_n^t)}^N_{n=1}),
\vspace{-1pt}
\end{equation}
where $\pi$ is the composition operator that aligns each actor based on their current pose $\bm{P}_n^t$, represents the set of poses for all actors, forming the complete 4D driving world at time $t$.
By iterating this process over time, we construct the dynamic 4D world that accounts for actor interplay, occlusions, and the evolving scene geometry. 

\subsection{Visual Scene Synthesis}
\label{sec:video}
\begin{figure}
    \centering
    \includegraphics[width=\linewidth]{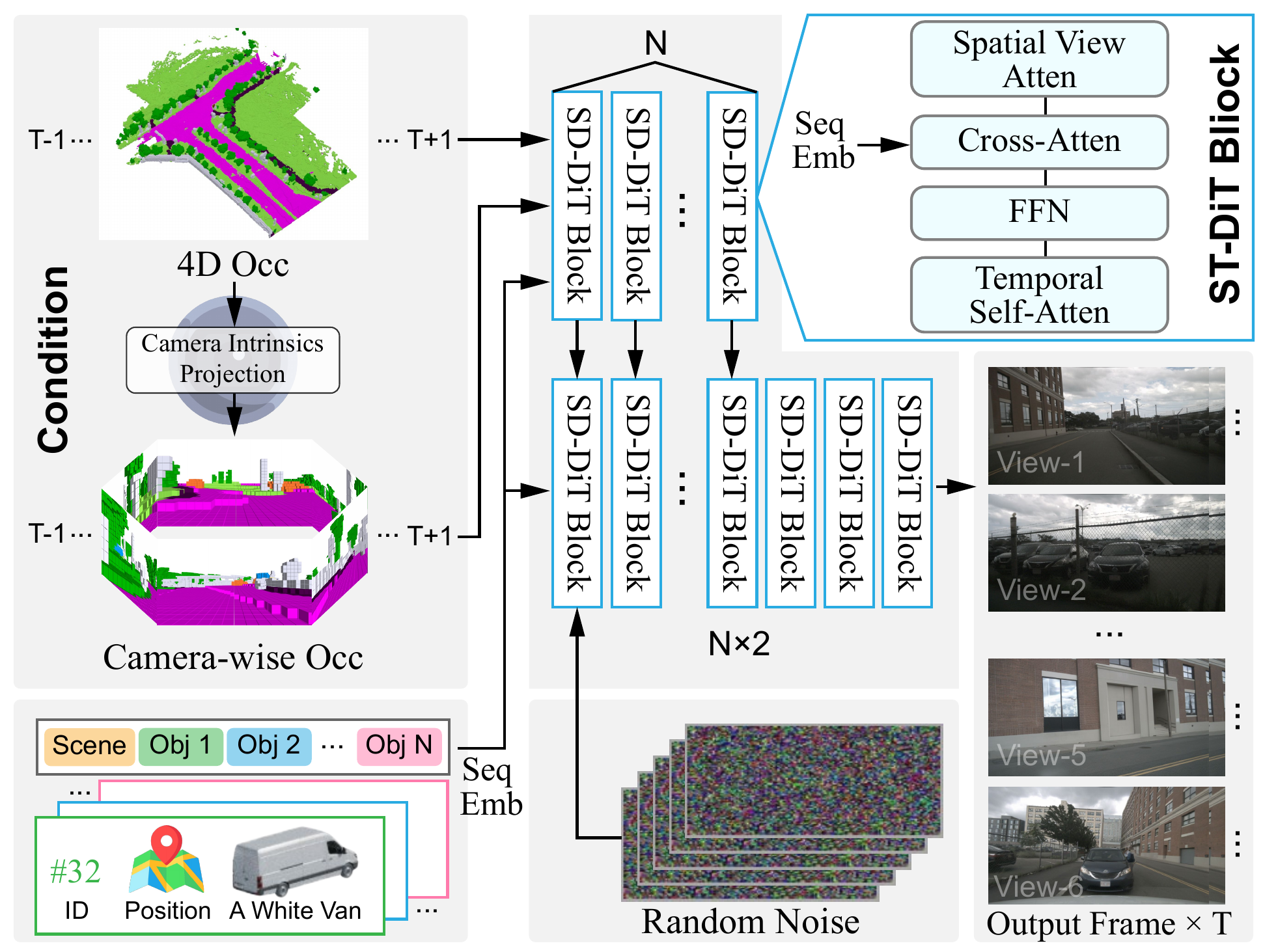}
    \caption{\textbf{Overview of the VideoDreamer.} The model conditions on 4D driving world and enriched actor embeddings (e.g., actor ID, position and caption). 
    This ensures high visual fidelity and geometric consistency in the generated driving simulations.}
    \label{fig:video}
\end{figure}

Previous generative models \cite{gao2023magicdrive, ma2024unleashing, wang2023drivedreamer, zhao2024drivedreamer2} tend to employ 2D visual conditions, which fail to accurately capture the geometric and semantic complexities inherent in real-world driving scenes. 
In response, our Visual Scene Synthesis employs VideoDreamer, a custom-designed video diffusion model, to transform the occupancy-driven 4D driving world constructed in the previous section into high-fidelity visual results.
The overall framework is shown in \cref{fig:video}.
Specifically, we introduce a Dual-path Condition Encoding strategy that focuses on encoding occupancy data as its primary condition. 
We further enhance the consistency of appearances across views and frames by developing an ID-aware Actor Encoding method. 
Finally, we integrate the Spatial-Temporal Diffusion Transformer (ST-DiT) from OpenSora~\cite{opensora} as the foundational technology to ensure visual consistency and generate artifact-free frames.

\noindent\textbf{Dual-path Condition Encoding.}
We design a dual-path condition encoding strategy aimed at effectively capturing occupancy data.
Given a set of driving world data $\bm{W}=\left\{ \bm{W}^{0},\bm{W}^{1},...,\bm{W}^{T} \right \}$ spanning frames from 0 up to $T$, we first encode them into a global feature $\bm{F}_{\text{global}} = \mathcal{F}^\text{4Docc}_\text{VAE}(\bm{W})$, capturing comprehensive geometric information and the spatial-temporal relationships within the scene.
Here, $\mathcal{F}^\text{4Docc}_\text{VAE}$ is the well-trained 4D encoder from occupancy VAE~\cite{wang2024occsora}, responsible for capturing global geometry of the scene.
This ensures that the overall structure and spatial layout of the scene are accurately represented.

Meanwhile, for $t^{th}$ frame (where $t=0,...,T$), 3D occupancy data $\bm{W}^t$ is projected into 2D semantic maps $\mathcal{M}_{v}^t$ for each view $v$ by camera extrinsics $\mathbf{T}^t$ and intrinsics $\mathbf{K}_v$:
\begin{equation}
\small
\vspace{-3pt}
    \mathcal{M}_{v}^t=\mathcal{Q}(\bm{W}^t,\mathbf{K}_v,\mathbf{T}^t),
\vspace{-1pt}
\end{equation}
where $\mathcal{Q}$ is the render and projection function~\cite{ramachandran2011mayavi}.
This process yields a sequence of semantic maps, $\mathcal{M} \in \mathbb{R}^{T \times 6 \times H \times W}$, with precise pixel-to-pixel alignment across views. Such alignment is essential for accurately capturing view-specific occlusions and depth variations.
$\mathcal{M}$ is then encoded into view-specific features $\bm{F}_\text{view}=\mathcal{F}_\text{VAE}^\text{Image}(\mathcal{M})$.
$\bm{F}_\text{view}$ and $\bm{F}_{\text{global}}$ are then integrated through cross-attention to obtain the unified representation $\bm{F}_\text{occ}$.

\noindent\textbf{ID-aware Actor Encoding.}
To ensure coherence of each actor’s appearance and identity information across the scene, we extract a fused sequence embedding $\bm{F}_\text{fuse}$:
\begin{equation}
\small
\vspace{-3pt}
    \bm{F}_\text{fuse}=\text{CONCAT}[\bm{F}_{\bm{W}},\bm{F}_{\bm{A}_{1}},...,\bm{F}_{\bm{A}_N}],
\vspace{-1pt}
\end{equation}
where $\bm{F}_{\bm{W}}$ is the embedding of the scene caption $\mathcal{L}_\text{scene}$, derived from the T5 encoder \cite{2020t5}. Each actor embedding $\bm{F}_{\bm{A}_{n}}$ is defined as:
\begin{equation}
\small
\vspace{-3pt}
    \bm{F}_{\bm{A}_n}=\text{CONCAT}[\mathcal{F}_\text{Fourier}(\bm{P}_n),\mathcal{F}_\text{Fourier}(\bm{I}_n),\mathcal{F}_{T5}(\bm{L}_{\bm{A}_i})].
\vspace{-1pt}
\end{equation}
Here, $\bm{P}_n$ represents the position of the $n$-th actor and $I_n$ is its unique ID. Both are encoded using Fourier encoding~\cite{mildenhall2021nerf}. The actor caption $\bm{L}_{\bm{A}_i}$ is encoded using the T5 \cite{2020t5}. 
Overall, by integrating geometry positions, textual descriptions, and identity information, we ensure coherent behavior and long-term visual consistency across frames.

\begin{figure}[t]
    \centering
    \includegraphics[width=\linewidth]{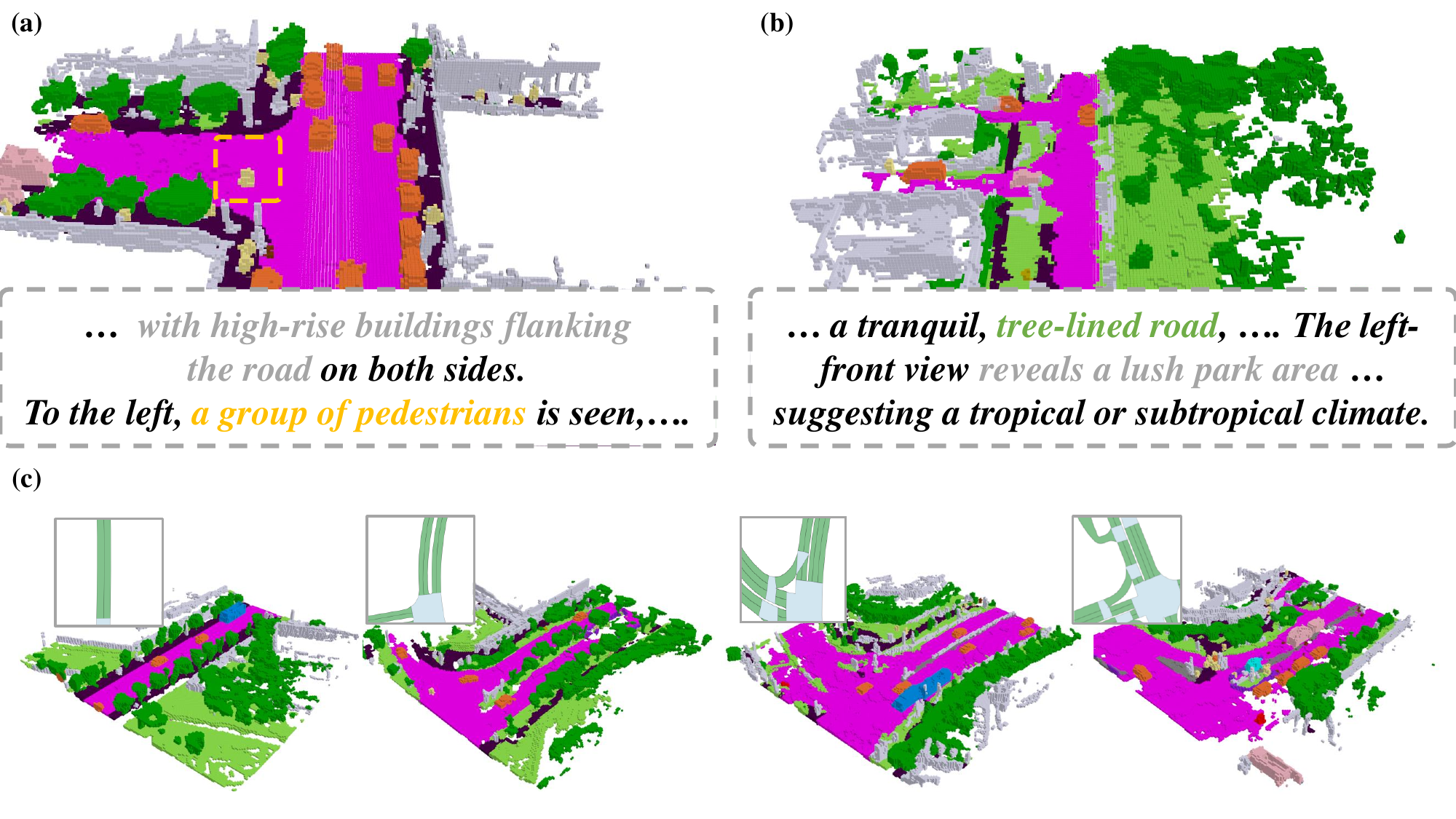}
    \caption{\small\textbf{Qualitative results of generated 3D scene}. OccDreamer uses text prompt and bev map as guidance to generate 3D scenes with controllable regional content and road structure.}
    \label{fig:vis_occ}
\end{figure}

\noindent\textbf{ControlNet-DiT.}
To enhance visual fidelity and temporal consistency in generated videos, we integrate the ST-DiT~\cite{opensora} as our denoiser $\epsilon^{v}_\theta$, utilizing ST-DiT blocks stacked with View-aware Spatial Self-Attention (VSSA), Temporal Self-Attention, Cross-Attention, and FFN.

In the video generation process, the input video sequence $\bm{V}\in\mathbb{R}^{v \times T \times H \times W}$ is first encoded into latent features $\bm{Z}^{\bm{V}} \in \mathbb{R}^{v \times T \times H' \times W' \times C}$. 
To ensure multi-view consistency while managing computational complexity, we modify the original Spatial self-attention~\cite{opensora} as VSSA. The latent feature $\bm{Z}^{\bm{V}}$ is reshaped from $v \times T \times H' \times W' \times C$ to $T \times (vH'W') \times C$, treating the combined view, height, and width dimensions $(vH'W')$ as the sequence length. This \textit{view-flatted} approach reduces the parameter overhead associated with cross-view attention \cite{gao2023magicdrive,wang2023drivingfuturemultiviewvisualdrivewm} while maintaining spatial coherence across different views. 
For external conditions, we apply a cross-attention layer that injects the fused sequence embedding $\bm{F}_\text{fuse}$ (derived from scene context and actor identities) as: $\bm{Z}^{\bm{V}}\!=\!\text{CrossAttention}(\bm{Z}^{\bm{V}},\bm{F}_\text{fuse}).$
Last but not least, we introduce ControlNet branch $\epsilon_\phi^{v}$ to ensure the occupancy control, inspired by PixArt-$\delta$~\cite{chen2024pixart}. This branch duplicates blocks from $\epsilon^{v}_\theta$ and takes $\bm{F}_\text{occ}$ as the input. The final diffusion process is:
\begin{equation}
        \small
        \vspace{-3pt}
        \mathcal{L}_\text{video}=\mathbb{E}_{\bm{F}_\text{occ},\bm{F}_\text{fuse},\bm{Z}^{\bm{V}},\epsilon^{v},\tau}[\left \| \epsilon^{v} - \epsilon^{v}_\theta (\bm{Z}^{\bm{V}}_\tau,\bm{F}_\text{fuse}, \tau,\epsilon_{\phi}^{v}(\bm{F}_\text{occ})) \right \|^2].
                \vspace{-1pt}
    \label{eq:difvideo}
\end{equation}
This approach ensures fine-grained control over spatial and temporal coherence, producing artifact-free frames that meet the high fidelity requirements for AD simulation.

\noindent\textbf{Auto-regressive Generation.} 
To generate temporally consistent and content-coherent video frames, we employ an auto-regressive generation strategy~\cite{opensora} based on first-$f$ frame mask. This approach models the dependency of each video frame on the previous frames, ensuring smooth transitions and maintaining visual continuity over time:
\[
\small
\vspace{-3pt}
\bm{V}^t = \mathcal{F}_\text{video}(\bm{V}^{t-1}, \bm{Z}^V, t),
\vspace{-1pt}
\]
where $\bm{V}^t$ is the video frame at time $t$, and $g_{\theta}$ is the video generation function that conditions on the latent representation $\bm{Z}^V$ and the previous frame $\bm{V}^{t-1}$. This ensures temporal coherence and infinite video sequences.

\subsection{Agent Interplay and Closed-Loop Simulation}
\label{sec:close-loop}
We enable seamless coordination of autonomous driving agents within the \ourmodel~simulation environment, categorizing agents into two primary types: the Ego Agent and Environment Agents.

\noindent\textbf{Ego Agent}: The Ego Agent represents the autonomous driving system under evaluation. Driven by end-to-end models~\cite{hu2023uniad,jiang2023vadvectorizedscenerepresentationvad,chen2024vadv2}, it receives visual input frames $\bm{V}^t$ and outputs the predicted control signal $\bm{c}_\text{ego}^{t+1}$ at each time $t$:
\begin{equation}
\small
    \bm{c}_\text{ego}^{t+1}=\mathcal{F}_\text{driving}(\bm{V}^t),
\end{equation}
where $\mathcal{F}_\text{driving}$ denotes the driving algorithm~\cite{hu2023uniad,chen2024vadv2,jiang2023vadvectorizedscenerepresentationvad}.

\begin{table*}[ht]
    \centering
    \setlength{\tabcolsep}{8pt}
    \resizebox{\textwidth}{!}{
    \begin{tabular}{c|c|cc|ccc|ccc}
    \hline
        \multirow{2}{*}{Source Data} & {Quality} &\multicolumn{2} {c|}{3DOD} &  \multicolumn{3} {c|}{Segmentation} & \multicolumn{3} {c}{$\text{L2}_{(\downarrow)}$}  \\
            \cline{2-10}
        &$\text{FVD}_{\small(\downarrow)}$ &{$\text{mAP}_{\small(\uparrow)}$}&$\text{NDS}_{(\uparrow)}$ & {$\text{Lanes}_{(\uparrow)}$} & $\text{Drivable}_{(\uparrow)}$ & {$\text{Divider}_{(\uparrow)}$}& {$\text{1.0s}$} & \text{2.0s} & \text{3.0s} \\
    \hline
    ori nuScenes\cite{Caesar_2020_CVPR_nuscenes} &—& 37.98 & 49.85 & 31.31 &69.14 & 25.93 & 0.51 & 0.98 & 1.65  \\
    MagicDrive~\cite{gao2023magicdrive} & 218.12 & 12.92 & 28.36 & 21.95 & 51.46 &17.10 &0.57 & 1.14 & 1.95  \\
        Panacea~\cite{wen2023panaceapanoramiccontrollablevideo} & 139 & 13.72 & 27.73 & 18.23 & 52.37 &17.21 &0.58 &1.14 &1.97  \\
    DriveArena~\cite{yang2024drivearena} & 185.32 & 16.06 & 30.03 & 26.14 & 59.37 &20.79 &0.56 &1.10 &1.89 \\
    \hline
    \ourmodel & \textbf{103.42} & 21.45 & 34.16 & 27.99 & 62.87 &22.29 &0.54 &1.10 &1.76    \\
    \hline
    \end{tabular}}
    \caption{\textbf{Quantitative comparison of video generation fidelity} for autonomous driving simulations on the nuScenes validation set. Using UniAD~\cite{hu2023uniad} for evaluation on generated data, \ourmodel~achieves high fidelity in video generation, closely matching real-world data across key perception and planning metrics.}
        \vspace{-3mm}
    \label{tab:sota_video}
\end{table*}

\noindent\textbf{Environment Agents}: Environment Agents are responsible for controlling the behavior and actions of all other actors within the simulation world $\bm{W}$. 
To achieve realistic info exchanges, we utilize the traffic flow engine~\cite{wenl2023limsim}, which supports multi-agent simulation. 
Environment Agents receive inputs from the simulation's state and output control signal $\bm{c}_\text{n}^{t+1}$ that dictate the movements and interplay of actors in the environment. Formally, this process can be described as:
\begin{equation}
\small
    \bm{c}_\text{n}^{t+1}=\mathcal{F}_\text{system}(\bm{W}^{t}),
\end{equation}
where $\mathcal{F}_\text{system}$ denotes the transportation simulator.

These agents take as input the visual frames $\bm{V}^t$ or previous state $\bm{W}^{t}$, and output actors' respective control signals $\bm{c}_\text{ego}^{t+1}$ and $\bm{c}_\text{n}^{t+1}$.
These signals influence the actor positions using \cref{actorposition} and further dynamiclly update the 4D driving world $\bm{W}^{t+1}$, maintaining the closed-loop simulation where the environment continuously adapts to the agents' actions.



%% file: sec/exp.tex
\section{Experiments}
\subsection{Settings}

\noindent\textbf{Dataset and Evaluation Metrics.}
We utilize real-world driving datasets nuScenes \cite{Caesar_2020_CVPR_nuscenes}. These datasets provide rich sensor data, including LiDAR point clouds, camera images, and high-definition maps, which are essential for generating accurate occupancy grids and semantic information.
We employ the following metrics to evaluate the performance of our simulation framework:
FVD and UniAD-based metrics~\cite{hu2023uniad} for visual quality and fidelity.
As for the evaluation of OccDreamer, we use FID and MMD~\cite{lee2024semcity}.
In open-loop evaluation, we borrow NC, DAC, TTC, and PDMS from NAVSIM~\cite{Dauner2024NEURIPS}. RC and ADS for the closed-loop following~\cite{yang2024drivearena}. Details are included in the supplementary.

\begin{figure*}[t]
    \centering
    \includegraphics[width=\linewidth]{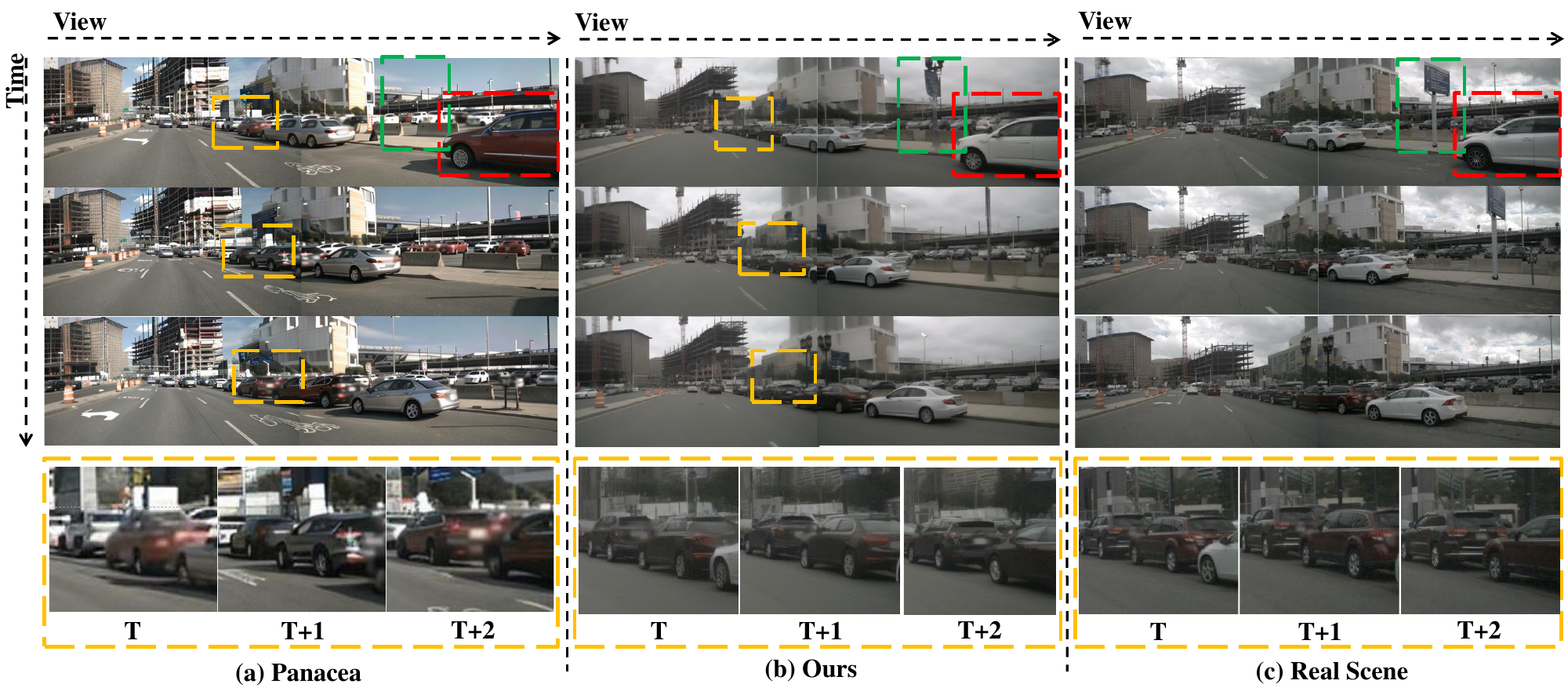}
        \vspace{-8mm}
    \caption{\textbf{Comparison of generated driving simulations from (a) Panacea~\cite{wen2023panaceapanoramiccontrollablevideo}, (b) Our method, and (c) Real Scene~\cite{Caesar_2020_CVPR_nuscenes} across multiple views and time steps.} \textcolor{red}{Red boxes} highlight instance-level appearance consistency, with our method maintaining more coherent actor appearances across frames.
    \textcolor{yellow}{Yellow boxes} illustrate temporal consistency, where our approach achieves smoother transitions over time. 
    \textcolor{green}{Green boxes} and background elements demonstrate our model's capability to capture and accurately model non-direct traffic participants, such as buildings and other environmental features, enhancing realism in the generated scene.
    }
    \label{fig:vis_video}
\end{figure*}

\noindent\textbf{Implement Details} are included in the supplementary.

\subsection{Visual Fidelity and Realism}
We first evaluate the fidelity of the voxel-based scenes from OccDreamer and video sequences from VideoDreamer against real nuScenes data to assess the domain gap.

\noindent\textbf{Scene Generation.}
We conduct quantitative and qualitative analyses to assess the fidelity of OccDreamer's generated occupancy data. Our method outperforms existing approaches quantitatively, as shown in \cref{tab:occ}. Specifically, SemCity employs a VAE trained on three planes to generate occupancy data while OccDreamer achieves pixel-to-pixel alignment based on BEV maps, resulting in more robust and accurate occupancy generation. 
\begin{table}[t]
    \centering
    \setlength{\tabcolsep}{25pt}
    \resizebox{\linewidth}{!}{
    \begin{tabular}{c|c|c}
    \hline
    {Methods}     &$\text{FID}_{\small(\downarrow)}$ &{$\text{MMD}_{\small(\downarrow)}$}  \\
    \hline
    Semcity~\cite{lee2024semcity} & 634 &0.251 \\
    \hline
    Ours & 274 & 0.082 \\
    \hline
    \end{tabular}}
    \caption{\textbf{Quantitative comparison of scene generation performance}, following the setup in SemCity. The scene data generated by our method is rendered into 2D images.}
    \label{tab:occ}
\end{table}condi
Qualitatively, \cref{fig:vis_occ} presents the results of OccDreamer with both caption and BEV map. To the best of our knowledge, our approach is the first to achieve text-guided occupancy generation. As demonstrated, abstract textual descriptions effectively guide the model to generate diverse scenes, including outdoor environments and urban landscapes. Under BEV map control, the generated samples accurately reflect road structures and lane configurations. These capabilities establish a solid foundation for the comprehensive modeling of the 4D driving world, ensuring that the generated scenes are both spatially continuous and logically consistent. More visualizations are provided in the supplementary materials.

\noindent\textbf{Video Generation.}
We further assess the performance of our video generation model, VideoDreamer, in generating coherent video sequences. 
We conduct a comparative analysis against leading models such as MagicDrive~\cite{gao2023magicdrive}, DriveArena~\cite{yang2024drivearena}, and Panacea~\cite{wen2023panaceapanoramiccontrollablevideo}. 
Using UniAD~\cite{hu2023uniad} as the evaluation framework, we assess the performance across crucial metrics including 3D object detection, BEV map segmentation, and trajectory planning. 
Our experimental results, summarized in \cref{tab:sota_video},  indicate that VideoDreamer significantly outperforms the baseline methods in these key metrics.
This performance underscores the advantages of occupancy data over traditional 2D conditions, affirming VideoDreamer's effectiveness in creating realistic, dynamic scenes that accurately simulate real-world driving complexities, supporting the development and validation of autonomous driving technologies.

Qualitatively, as shown in \cref{fig:vis_video}, 
our approach demonstrates markedly superior performance in maintaining coherence and fidelity over time, significantly outperforming others. 
This success is primarily due to the geometry-aware occupancy encoding and instance encoding that ensures appearance consistency, highlighting VideoDreamer's robustness in creating visually consistent, detailed driving environments that capture real-world scene dynamics. More visualizations are included in the supplementary materials.

\subsection{Open-loop Evaluation}
In the open-loop setting, where AD agents passively receive environment inputs without influencing the simulation dynamics, we assess the performance of UniAD~\cite{hu2023uniad} on \ourmodel~and alternative simulators. 
Using 100 predefined trajectories, \cref{tab:openloop} demonstrates superior PDMS scores achieved by AD agents on \ourmodel, suggesting that its higher visual fidelity reduces perceptual inaccuracies that can misguide agent decisions. This improvement highlights \ourmodel’s suitability for testing AD systems in environments that closely emulate real-world conditions.
\begin{table}[t]
    \centering
    \setlength{\tabcolsep}{6pt}
    \resizebox{\linewidth}{!}{
    \begin{tabular}{c|cccc}
    \hline
    {Data Source}     &$\text{NC}_{\small(\uparrow)}$ &{$\text{DAC}_{\small(\uparrow)}$} & {$\text{TTC}_{\small(\uparrow)}$}&$\text{PDMS}_{\small(\uparrow)}$  \\
    \hline
    nuScenes~\cite{Caesar_2020_CVPR_nuscenes} & 0.993& 0.995 & 0.947 &0.910 \\
    DriveArena~\cite{yang2024drivearena} & 0.792 &0.942 &0.771 &0.636 \\
    \hline
        DriveArena*~\cite{yang2024drivearena} & 0.829 &0.964 &0.812 &0.698 \\
    \ourmodel & 0.852 & 0.968 &0.853 & 0.742\\
    \hline
    \end{tabular}}
    \caption{\textbf{Open-loop evaluation of UniAD~\cite{hu2023uniad}} on original nuScenes sequences, DriveArena's reported results, and our pre-defined 100-trajectory test for DriveArena* and \ourmodel.}
    \label{tab:openloop}
        \vspace{-5mm}
\end{table}
\subsection{Closed-loop Evaluation}
In the closed-loop evaluation, AD agents receive visual input and output control signals, thus interactively shaping the simulation. 
This evaluation setup involves 100 pre-defined trajectories to test across controlled yet varied scenarios. 
As shown in \cref{tab:closedloop}, UniAD~\cite{hu2023uniad} exhibits relatively low Route Completion (RC) scores, with completing an average of only 11.7\% of each route. 
Compared to DriveArena~\cite{yang2024drivearena}, our simulation yields consistently better performance metrics, and, along with the superior visual fidelity demonstrated in~\cref{tab:sota_video}, affirms \ourmodel’s effectiveness as a high-fidelity platform for AD algorithm validation.

\begin{table}[t]
    \centering
    \setlength{\tabcolsep}{12pt}
    \resizebox{\linewidth}{!}{
    \begin{tabular}{c|ccc}
    \hline
    {Route}     &$\text{PDMS}_{\small(\uparrow)}$ &{$\text{RC}_{\small(\uparrow)}$} &$\text{ADS}_{\small(\uparrow)}$  \\
    \hline
    DriveArena~\cite{yang2024drivearena} & 0.6901 &0.0641 &0.0508 \\
    \hline
    \ourmodel & 0.7281 & 0.1170 &0.0851\\
    \hline
    \end{tabular}}
    \caption{\textbf{Closed-loop evaluation of UniAD's~\cite{hu2023uniad} performance} on 100 pre-defined routes.}
    \label{tab:closedloop}
        \vspace{-3mm}
\end{table}

\section{Conclusion}
We presented \ourmodel, a generative closed-loop simulation framework that bridges the gap between traditional closed-loop simulations and open-loop generative models. Through advanced occupancy-based modeling and controllable generation mechanisms, \ourmodel~creates realistic, high-fidelity simulation for autonomous driving. Our experiments demonstrate superior visual quality, temporal consistency, and the ability to effectively test AD algorithms in dynamic environments. In future work, we aim to enhance dynamic agent behaviors and incorporate more complex real-world scenarios to further improve the robustness of our simulations.